\documentclass[letterpaper, 10 pt, conference]{ieeeconf}
\usepackage{graphicx}
\usepackage{grffile}
\usepackage[font=small,labelfont=bf]{caption}
\usepackage{amsmath}
\usepackage[pdftex,pdftitle={Your Title},pdfversion=1.5]{hyperref}
\IEEEoverridecommandlockouts 
\usepackage[T1]{fontenc}
\usepackage{makecell}
\usepackage{array}
\usepackage{multirow}
\usepackage{threeparttable}
\usepackage{algorithm}
\usepackage{algpseudocode}
\usepackage{makecell}

\usepackage{graphics}           
\usepackage{times}              
\usepackage{amsmath}            
\usepackage{amssymb}            
\usepackage{graphicx}
\usepackage{algorithm}
\usepackage{algpseudocode}
\usepackage{wrapfig}
\usepackage{booktabs}
\usepackage{color}
\definecolor{instructioncolor}{rgb}{.5,.5,.5}

\usepackage[font=small]{caption}

\def\figref#1{Fig.~\ref{#1}}

\def\eqref#1{Eq.~(\ref{#1})}

\makeatletter
\usepackage{xspace}
\DeclareRobustCommand\onedot{\futurelet\@let@token\@onedot}
\def\@onedot{\ifx\@let@token.\else.\null\fi\xspace}

\def\etal{{et al}\onedot}
\makeatother

\usepackage{array}
\newcolumntype{L}[1]{>{\raggedright\let\newline\\\arraybackslash\hspace{0pt}}m{#1}}
\newcolumntype{C}[1]{>{\centering\let\newline\\\arraybackslash\hspace{0pt}}m{#1}}
\newcolumntype{R}[1]{>{\raggedleft\let\newline\\\arraybackslash\hspace{0pt}}m{#1}}

\title{\LARGE \bf NuExo: A Wearable Exoskeleton Covering all Upper Limb ROM for Outdoor Data Collection and Teleoperation of Humanoid Robots}

\author{Rui\,Zhong\textsuperscript{ \#},~Chuang\,Cheng\textsuperscript{ \#},~Junpeng\,Xu,~Yantong\,Wei,~Ce\,Guo,~Daoxun\,Zhang,~Wei\,Dai\,$^*$,~Huimin\,Lu\,$^*$
\vspace{-1.2cm}
  \thanks{All authors are with the College of Intelligence Science and Technology, and the National Key Laboratory of Equipment State Sensing and Smart Support, National University of Defense Technology, Changsha, China.} 
    \thanks{\text{\#} These authors contributed equally to this work and share co-first authorship.}
  \thanks{This work was supported in part by the National Science Foundation of China under Grant 62203460, U22A2059, and 62403478, Young Elite Scientists Sponsorship Program by CAST (No. 2023QNRC001), as well as the Innovation Science Foundation of National University of Defense Technology under Grant 24-ZZCX-GZZ-11. ($^*$\,Corresponding authors: Huimin Lu, Wei Dai.)
  }%
}
\begin{document}
\maketitle
\thispagestyle{empty}
\pagestyle{empty}

\begin{abstract}
The evolution from motion capture and teleoperation to robot skill learning has emerged as a hotspot and critical pathway for advancing embodied intelligence. 
However, existing systems still face a persistent gap in simultaneously achieving four objectives:
accurate tracking of full upper limb movements over extended durations~(Accuracy),
ergonomic adaptation to human biomechanics~(Comfort),
versatile data collection (e.g., force data) and compatibility with humanoid robots~(Versatility), 
and lightweight design for outdoor daily use~(Convenience).
We present a wearable exoskeleton system, incorporating user-friendly immersive teleoperation and multi-modal sensing collection to bridge this gap. Due to the features of a novel shoulder mechanism with synchronized linkage and timing belt transmission, this system can adapt well to compound shoulder movements and replicate 100\% coverage of natural upper limb motion ranges. Weighing 5.2\,kg, NuExo supports backpack-type use and can be conveniently applied in daily outdoor scenarios. Furthermore, we develop a unified intuitive teleoperation framework and a comprehensive data collection system integrating multi-modal sensing for various humanoid robots.
Experiments across distinct humanoid platforms and different users validate our exoskeleton's superiority in motion range and flexibility, while confirming its stability in data collection and teleoperation accuracy in dynamic scenarios. 

\end{abstract}
\section{INTRODUCTION}
\label{sec:intro}

In recent years, humanoid robot technology and imitation reinforcement learning techniques have seen significant development~\cite{Zhao2023rss_aloha,fu2024humanplus}. Imitation reinforcement learning provides a viable pathway for skill acquisition in humanoid robots~\cite{chi2024ijrr_DiffusionPolicy}.
Collecting human and robot motion data is a crucial foundation for this technology.
The current data collection methods for robot actions mainly rely on teleoperation or motion capture, including:
hand controllers as teleoperation devices~\cite{Zhao2023rss_aloha,Wu2024irosGELLO};
user-friendly handheld and glove-based control collection devices~\cite{chi2024rss_UMI,wang2024arxiv_DexCap};
immersive control devices such as VR~\cite{Zhang2018icra_deepImitationVR} and XR~\cite{Ding2024arxiv_BunnyVisionPro};
high-precision vision motion capture devices~\cite{qin2023rss_anyteleop} and inertial motion capture devices~\cite{cheng2024corl_Open-TeleVision}.

These motion data collection and teleoperation devices have their respective advantages, but they also face the following limitations in various degrees: 1)~a trade-off between the accuracy of collected data and the constraints of application scenarios; 2)~the conflict between ease of use and data richness; 3)~the limitations in terms of human adaptability, operational intuitiveness, and the versatility of operational objects.
Therefore, there is still a lack of unified teleoperation and motion data collection devices that can simultaneously achieve the following features: intuitive operation without the need for training, general compatibility with different humanoid robots, the ability to handle various data types (especially force data), data stability without drift over time, and practical portability for deployment in unstructured outdoor environments and daily living scenarios.

\begin{figure}[t]
  \centering
\includegraphics[width=1\linewidth]{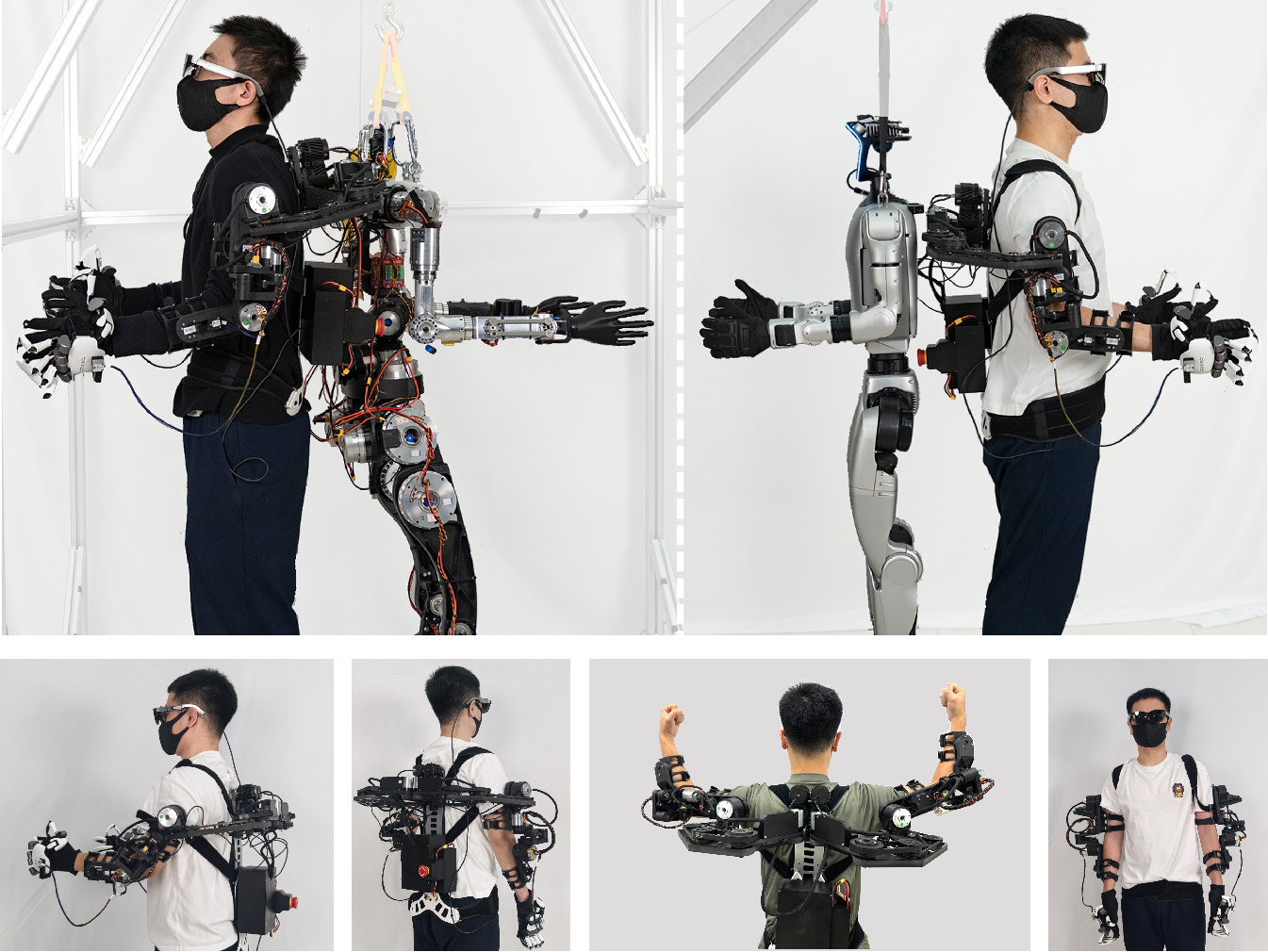}
  \caption{NuExo: A backpack-mounted active-joint humanoid robot exoskeleton teleoperation system with large motion range.}
  \label{fig:main}
  \vspace{-0.7cm}
\end{figure}

In response to these urgent requirements, the contributions of this paper are as follows:

\begin{itemize}
 \item We designed a \textbf{wearable, human anatomy-conformed, motor-driven exoskeleton} with a special shoulder structure maintaining full upper-limb motion coverage. This first wearable implementation of sternoclavicular compensation resolves motor interference issues, setting a new benchmark for anatomical exoskeletons.
\item We proposed a \textbf{unified teleoperation system integrating multi-modal sensing} (force-torque, joint position, egocentric vision) and a calibration-free control framework. The system enables zero-shot operation of humanoid platforms with high manipulation accuracy in complex tasks. 
\item We implemented the teleoperation framework and collection software in \textbf{different humanoid platforms}. Our experiments show the system's versatility and strong manipulation performance with various operators and robots. Additionally, the range of motion~(ROM) measuring and the collection data accuracy comparison experiment verified our exoskeleton's adaptability with humans, and the accuracy of long-term data collection.
\end{itemize}

\section{RELATED WORK}
\label{sec:related_work}
\vspace{-0.1cm}
\subsection{Driven by the Demand for Skill Imitation Learning}
\vspace{-0.1cm}
In recent years, imitation learning has attracted considerable attention from researchers~\cite{chi2024ijrr_DiffusionPolicy,chi2024rss_UMI,fu2024corl_MobileAloha}. 
However, skill learning in existing robotic systems is still constrained by the dataset's scale and diversity~\cite{Wu2024irosGELLO}. 
As a result, many researchers have begun to innovate data collection methods and devices.
Zhao et al.~\cite{Zhao2023rss_aloha} proposed a framework for collecting homogeneous low-cost teleoperation data, attracting significant interest. 
However, because it currently focuses primarily on algorithms, the teleoperation device is designed to be the same as the controlled manipulator. 
Its second-generation version, Mobile Aloha~\cite{fu2024corl_MobileAloha}, expands its application to indoor mobile scenarios. At the same time, Wu et al. have proposed a similar improvement device named Gello~\cite{Wu2024irosGELLO} to extend it to general industrial manipulator applications.

In addition, a lightweight handheld data collection device UMI has been introduced by~\cite{chi2024rss_UMI}. 
It offers excellent usability and employs a heterogeneous data transmission model, which makes it suitable for general industrial manipulators~\cite{wu2024arxiv_Fast-UMI} as well as manipulators mounted on quadruped robots~\cite{Ha2024arxivUMIonLeg}. 
This device has improved in~\cite{Liu2024arxivForceMimicFI} named Forcemimic by adding a force sensor, enabling it to perform force-controlled skill transmission. 
However, this method currently applies only to the manipulators equipped with two-finger grippers. It implements the heterogeneous operating modes, mainly allowing the robot to learn the implicit relationship between the end effector and the world; however, this method makes it difficult for the robot to acquire a comprehensive arm movement strategy. 

To extend the operation of the two-finger gripper to dexterous hand operations, a work named BennyVisionPro~\cite{Ding2024arxiv_BunnyVisionPro} has open-sourced a remote operation method for the dual-arm robot with a dexterous robotic hand by using Apple$^\circledR$ Vision Pro based on mixed reality~(MR).

In addition, similar to MR, motion capture technology also provides alternative pathways for implementing teleoperation and imitation learning. Inertial motion capture collects information about the state of human motion through relationships between multiple IMUs on the body, allowing for the collection of human motion data. However, issues such as drift over prolonged use can affect the accuracy and precision of data collection. Visual motion capture with markers offers good accuracy~\cite{qin2023rss_anyteleop}. However, it is often limited by usage scenarios and costs, making it expensive, typically restricted to fixed indoor spaces, and whose marker deployment also requires time. Directly using a small number of cameras for motion capture based on skeletal and joint visual recognition can solve this problem. 
Methods that integrate action recognition, manipulation, and learning have been proposed and implemented by Anyteleope~\cite{qin2023rss_anyteleop} and Humanoidplus~\cite{fu2024humanplus}. Humanoidplus is a teleoperation method for the whole humanoid robot body, the operation accuracy is limited by the recognition algorithms.  

The development of exoskeleton technology provides new ideas for this research. The joint data it collects typically comes from motor encoders, which are more stable, and the use of such devices aligns better with human movement intuition. 
AirExo innovatively transformed homogeneous operational devices~\cite{fang2024icra_airexo} such as Aloha~\cite{Zhao2023rss_aloha} and Gello~\cite{Wu2024irosGELLO} into a wearable exoskeleton form, achieving imitation learning for several industrial manipulators. However, the entire system is built using encoders and similarly lacks force data collection.
 
\vspace{-0.1cm}
\subsection{Inspiration from Upper Limb Rehabilitation Exoskeletons
}
\vspace{-0.1cm}

Upper limb exoskeletons have significantly improved rehabilitation robotics over the past decade. The seminal work by~\cite{kim2017ijrr} introduced the exoskeleton Harmony-SHR, which pioneered shoulder anatomical motion compensation to enhance movement compliance and coordination during rehabilitation. This breakthrough established shoulder joint compensation and compound motion synchronization as critical design priorities, driving subsequent innovations to expand the operational workspace and improve motion smoothness.

Recent efforts have extended these principles to teleoperation systems\cite{zimmermann2023tro, Pan2022ral, Cheng2024tmech}. Cheng~\etal proposed a 9-DoF exoskeleton with dynamic compensation for high-speed manipulation and homo-hetero teleoperation mapping. While achieving full upper limb kinematic coverage, their cockpit-based design inherently limits portability and outdoor applicability. These developments demonstrate that anatomical shoulder compensation improves teleoperation transparency and user experience. However, the prevailing approach of adding compensatory actuators (typically two extra motors) results in over-engineered shoulder modules with up to five actuators. This creates critical challenges in mechanical interference prevention and weight distribution, severely constraining the lightweight adoption of anatomically inspired designs in field-deployable systems.

\begin{figure*}[t]
  \centering
  \includegraphics[width=1\linewidth]{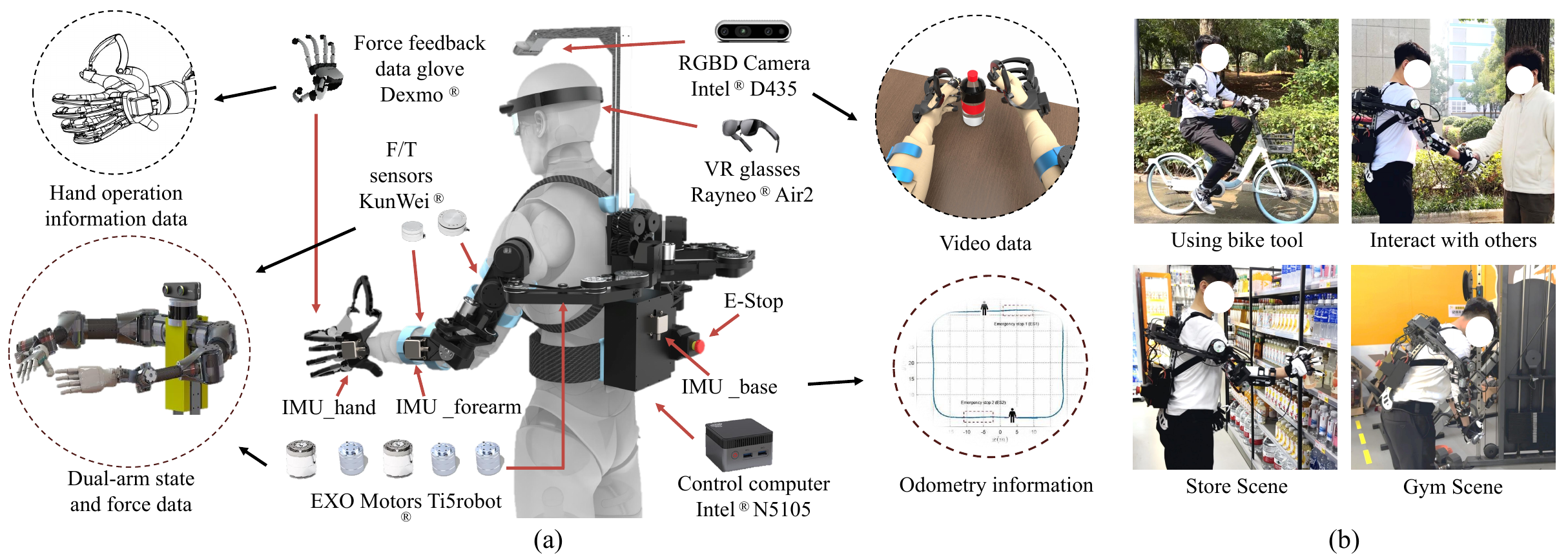}
  \caption{Overview of the NuExo teleoperation exoskeleton data collection system (a) and the NuExo in daily scene (b).}
  \label{fig:system}
\end{figure*}

Therefore, this paper proposes an innovative linkage-timing belt synergetic shoulder mechanism that breaks the constraints of conventional exoskeletal shoulder designs. Our approach resolves persistent challenges in actuator placement and mechanical interference by eliminating one actuator while maintaining equivalent active motion compensation performance. This breakthrough fully covers natural upper limb motion ranges and pioneers the first lightweight backpack-mounted implementation of anatomically compensated exoskeletons. Based on this, we develop a unified immersive teleoperation system with integrated multi-modal sensing for data collection for humanoid robots. It aims to provide a convenient, versatile, comfortable, precise, and comprehensive hardware solution for teleoperation and data collection in humanoid robots.
\section{OVERVIEW OF THE TELEOPERATION EXOSKELETON DATA COLLECTION SYSTEM}
The overview of the NuExo teleoperation and data collection exoskeleton system is shown in~\figref{fig:system}, which can be divided into the following major components: the upper limb exoskeleton main body, the manipulated humanoid robot, the visual display and acquisition module, and the control and battery unit.

\textbf{Exoskeleton Design:}
As the core of the system, the upper limb exoskeleton integrates shoulder, elbow, wrist, and hand-mounted data glove modules. The shoulder module employs an innovative linkage and timing belt mechanism that reduces the number of actuators while compensating for sternoclavicular joint motion. This biomimetic design aligns with anatomical principles of human upper limb movement, ensuring natural shoulder mobility for users and resolving actuator placement challenges in shoulder-mounted exoskeletons for portable systems.
The elbow module adopts a dual-actuator structure optimized to address mechanical constraints during horizontal rotation, with detailed design and kinematic principles elaborated in Sections 4.1,~4.2. The total weight of the exoskeleton is 5.2\,kg and is made mainly of carbon fiber plates and 3D printed PLA plastic components.

The sky-blue parts of the arm in the model in \figref{fig:system} (a) denote binding interfaces housing six-axis force sensors, which connect users to the exoskeleton. These sensors collect upper limb kinetic data during operation—a feature absent in most existing systems—and synergize with our collaborative control algorithm (Section 4.3) to enable floating assistive motion, significantly enhancing operational flexibility.
For the wrist module, a differential IMU configuration replaces conventional mechanisms to prioritize ergonomic compatibility, simplifying integration with force-feedback data gloves. An optional RGB-D camera mounted on the exoskeleton captures first-person-view upper limb motion videos, aligning with human visual perception.

\textbf{Teleoperation and Data Acquisition:}
To ensure cross-platform adaptability for diverse humanoid robots, we developed a universal control framework (Section 4.3). A lightweight AR headset provides real-time display of the robot’s first-person perspective, enhancing operational immersion. For large-scale outdoor motion data collection, an IMU-based odometry system is integrated into the main controller, enabling spatiotemporal synchronization of motion and positional data.
This system uniquely supports simultaneous acquisition of: 
(1)~robotic teleoperation data, (2)~human upper limb kinematic data, (3)~first-person-view motion videos, (4)~finger motion data, (5)~odometry data, (6)~upper limb force feedback.
To our knowledge, this represents \textbf{the most comprehensive multimodal data collection platform for humanoid robot upper limb studies}, with outdoor mobility further expanding its spatial and temporal scalability. \figref{fig:system} (b) presents the user wearing the exoskeleton in the daily outdoor scene, whose motions are influenced little.

\section{INNOVATIVE DESIGN OF EXOSKELETON STRUCTURE AND TELEOPERATION CONTROL}
\label{sec:Design of the exoskeleton}
\begin{figure*}[t]
  \centering
  \includegraphics[width=0.91\linewidth]{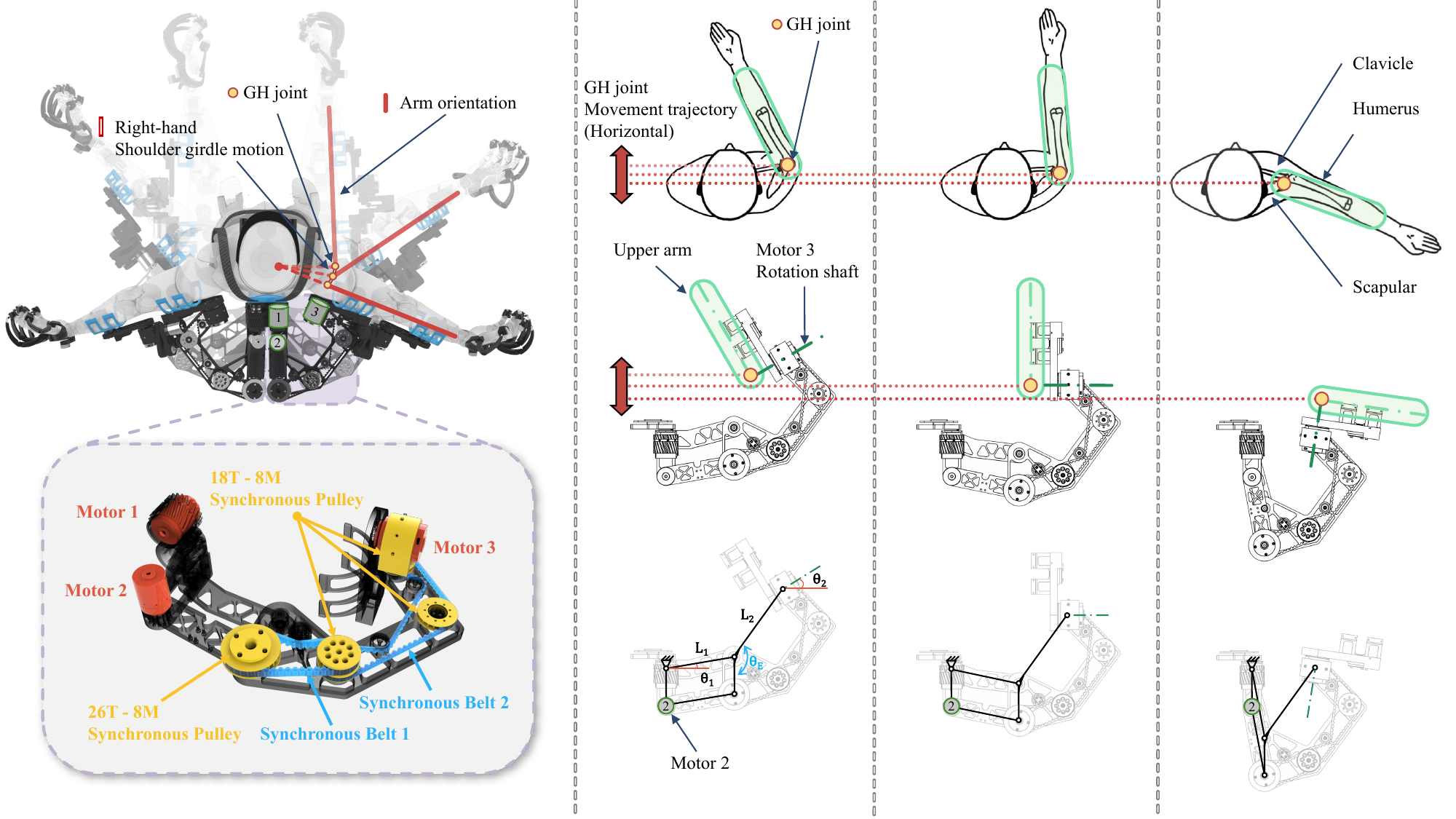}
  \caption{The schematic representation of the shoulder mechanical structure of the exoskeleton tracking the dynamically changing center of GH joint in the horizontal plane.}
  \label{fig:shoulder movement}
  \vspace{-0.4cm}
\end{figure*}

\begin{figure*}[!t]
  \centering
  \includegraphics[width=0.91\linewidth]{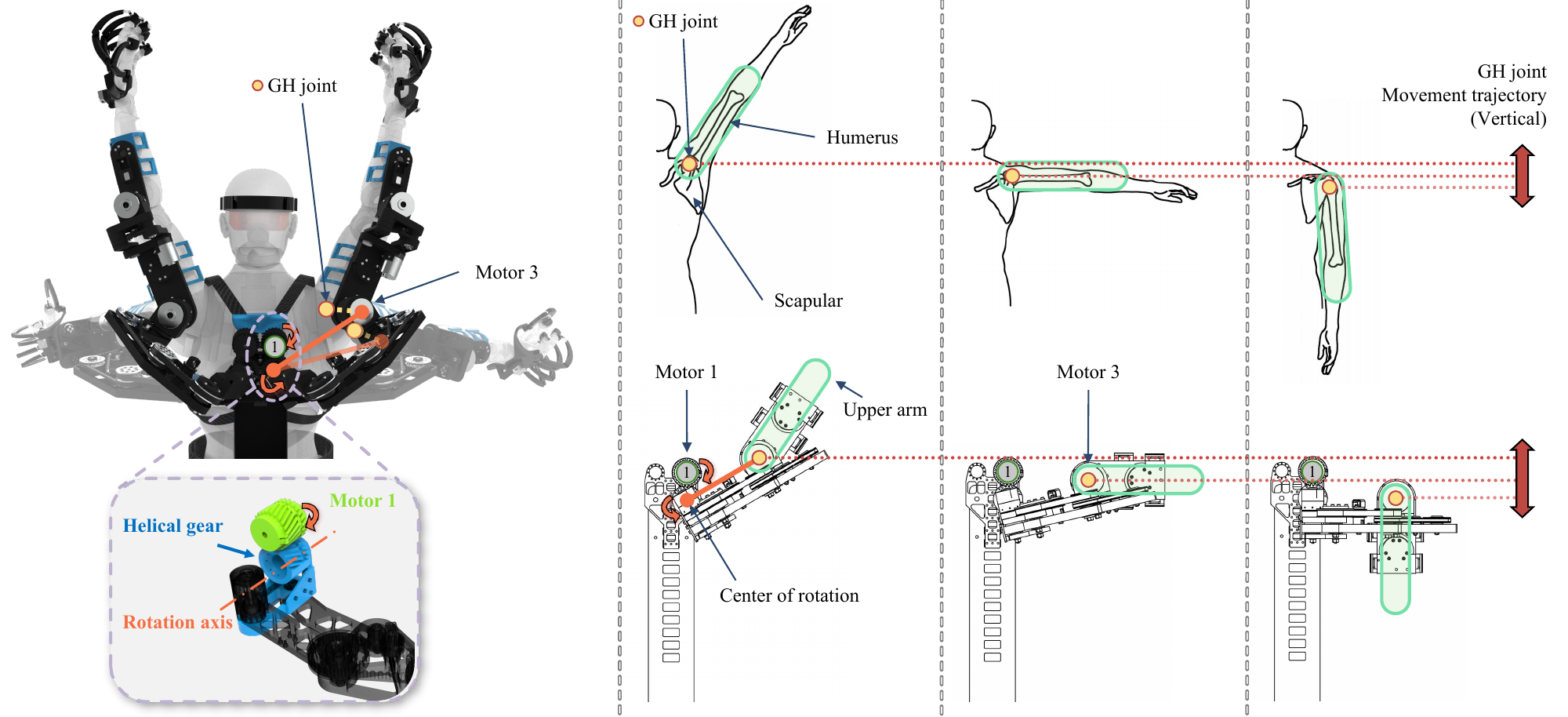}
  \caption{The schematic illustration of how the mechanical structure tracks the dynamically changing center of GH joint in the vertical direction.}
  \label{fig:back movement}
  \vspace{-0.5cm}
\end{figure*}

\subsection{The Shoulder Structure Design of the Exoskeleton}
The human shoulder can be conceptualized as a spherical hinge structure with three degrees of freedom. The geometric center of this spherical hinge is commonly regarded as the Glenohumeral (GH) joint. Nevertheless, during natural human movement, this geometric center is not stationary relative to the body but rather exhibits a certain degree of movement freedom\cite{kim2017ijrr}. Most exoskeleton systems have neglected this aspect, leading to a misalignment between the shoulder movement center of the exoskeleton and GH joint. The Harmony Exo~\cite{kim2017ijrr} exoskeleton system and the ANYexo~\cite{zimmermann2023tro} exoskeleton system consider the DOF of GH joint as two specific types: horizontal movement (forward and backward) and vertical elevation. They have made corresponding structural compensations for these movements, thereby improving the comfort of wearing of the exoskeleton.

Our findings indicate that the horizontal movement of the GH joint is closely associated with the movement of the upper arm. When the human upper arm rotates naturally without deliberately moving the shoulder, GH joint will also move accordingly. This observation suggests that one DOF of the shoulder spherical hinge and the DOF of the forward (or backward) movement of GH joint are not entirely independent, but rather have a mapping relationship. Consequently, we can utilize a mechanical transmission structure to combine these two degrees of freedom into one. By doing so, we can reduce the number of driving motors without sacrificing the tracking performance, ultimately decreasing the weight of the exoskeleton. \figref{fig:shoulder movement} illustrates the tracking performance of our exoskeleton shoulder during horizontal adduction and abduction movements of the upper arm. These mechanical parameters are shown in Tab.~\ref{tab:Exo Shoulder}.

\begin{figure*}[t]
  \centering
   \includegraphics[width=1\linewidth]{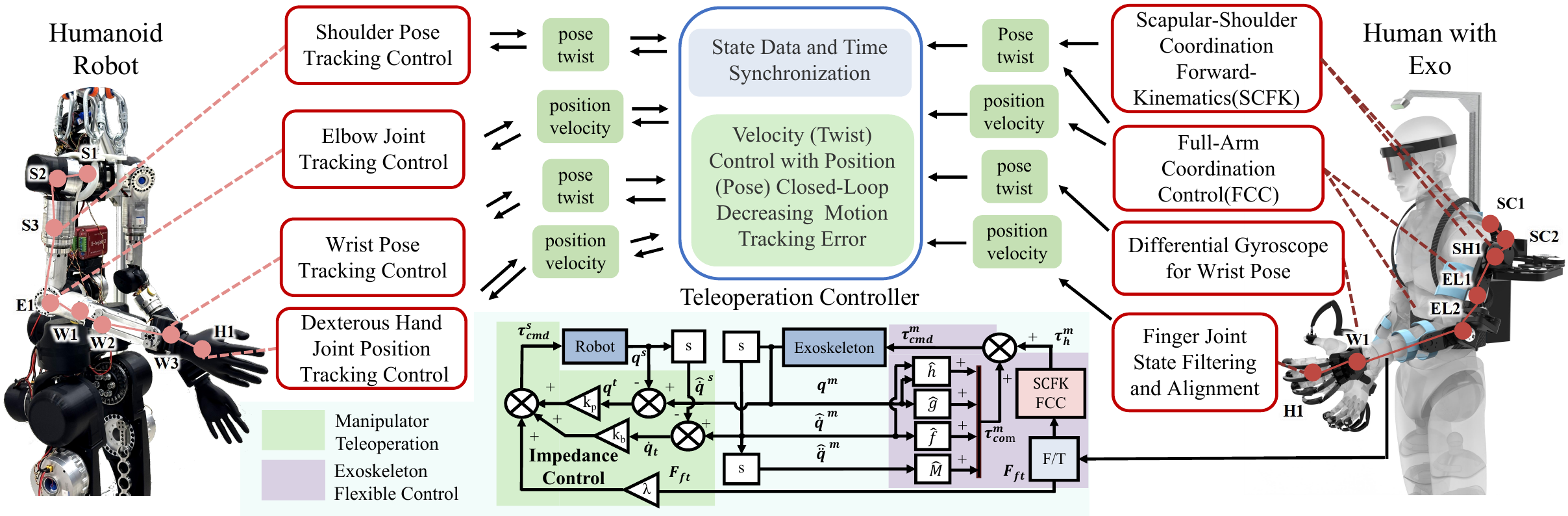}
    \captionof{figure}{Control diagram of the Teleoperation. 
     the shoulder and wrist posture~(the elbow and hand joints position) of the human arm with the exoskeleton~$\boldsymbol{q}^{m}$~and that of the humanoid robot~$\boldsymbol{q}^{s}$, and the angular velocity and joints rotate of the both ${\hat{\dot{\boldsymbol{q}}}}^{m}$, ${\hat{\dot{\boldsymbol{q}}}}^{s}$. They are transmitted to the teleoperation controller. The differences between~$\boldsymbol{q}^{m}$,~$\boldsymbol{q}^{s}$~and~${\hat{\dot{\boldsymbol{q}}}}^{m}$,${\hat{\dot{\boldsymbol{q}}}}^{s}$~form the control target input~$\boldsymbol{q}_{t}$,~${\dot{\boldsymbol{q}}}_{t}$ to the impedance controller, which outputs the torque control varibles ${\boldsymbol{\tau}}_{cmd}^{s}$ for each component of the controlled robot. Regarding Exo. control, we implement our previous work the binding alignment strategy~(BAS) and the full-arm coordination mechanism~(FCM)~\cite{cheng2025flexibleexoskeletoncontrolbased}to produce the coordination control variable~$\boldsymbol{\tau}_{h}^{m}$~\cite{Cheng2024tmech}, which along with the dynamics compensation ~$\boldsymbol{\tau}_{com}$~(consist of inertia~$\hat{\boldsymbol{M}}$, Coriolis force~$\hat{\boldsymbol{h}}$, gravity~$\hat{\boldsymbol{g}}$, and friction compensation~$\hat{\boldsymbol{f}}$), constitutes the control variable of the exoskeleton motors~$\boldsymbol{\tau}_{cmd}^{m}$.
     }
     
    \label{fig:exo_control}
    \vspace{-0.5cm}
\end{figure*}

\begin{table}[h]
    \centering
    \caption{The Main Parameters of Shoulder Structure in \figref{fig:shoulder movement}.}
    \setlength{\tabcolsep}{10pt}
    \begin{tabular}{cccccc}
        \toprule
        \multicolumn{2}{c}{$L_1$:~150\,mm} & \multicolumn{2}{c}{$L_2$:~187\,mm} & \multicolumn{2}{c}{$\theta_E$:~2.508\,rad}  \\
        \midrule
         \multicolumn{3}{c}{$\theta_1$:~Rotation Angle of Motor 2}& \multicolumn{3}{c}{$\theta_2$:~1.444$\theta_1$~+~0.938\,rad}\\
        \bottomrule
    \end{tabular}
 \label{tab:Exo Shoulder}
 \vspace{-0.5cm}
\end{table}

In addition to the protraction–retraction movement, GH joint also undergoes elevation–depression movement when the arm is raised. To compensate for this movement, we have designed the structure depicted in \figref{fig:back movement}. This structure allows the exoskeleton to adapt to the dynamic changes in the vertical position of GH joint.

\subsection{Design of the Elbow Structure}

\begin{figure}[h]
  \centering
  \includegraphics[width=0.9\linewidth]{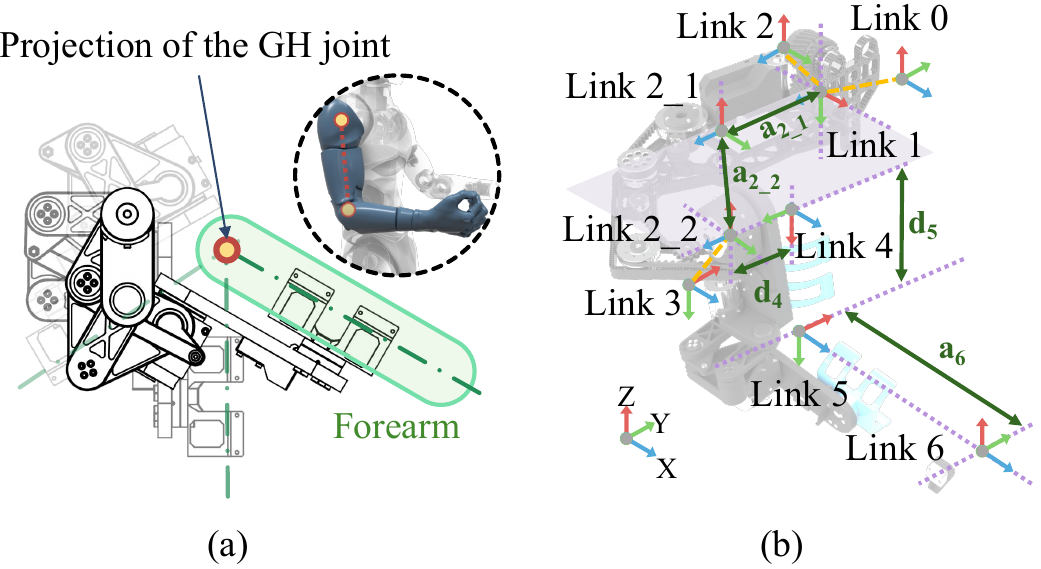}
  \caption{(a): Elbow design. (b): Definition of the exoskeleton coordinate system}
  \label{fig:Exo elbow}
\vspace{-0.5cm}
\end{figure}

We reposition a motor of the shoulder joint to the elbow~\cite{Cheng2024tmech} (illustrated in \figref{fig:Exo elbow} (a)). This strategic relocation effectively eliminates the potential interference between the motor and the human head during arm elevation, a common issue in many conventional designs. This innovative approach enables us to utilize only one motor on the shoulder. In contrast, traditional exoskeleton systems equipped with three actively driven degrees of freedom in the shoulder lack such a streamlined and efficient design. Through this distinctive shoulder design, our exoskeleton attains an impressively wide ROM. The DH parameters table of NuExo is shown in the Tab.~\ref{tab:DH}. The joints 2-1, 2-2 and 3 are the passive DOF of the linkage system. The rotational corresponding relationship between Link2 and 2\_1, Link2\_1 and 2\_2, Link2\_2 and 3 satisfy the geometric relationship in Tab.~\ref{tab:Exo Shoulder}.

\begin{table}[tbp]
\caption{DH-Parameters of the Exo.}
    \centering
    \begin{tabular}{lrrrr}
      \toprule
       Link & $\theta$ & d  & a & $\alpha$ \\
        \midrule
        0~-~1       & $-\pi/2$        & 0       & 0         & $-\pi/2$ \\
        1~-~2       & 0               & 0       &0          & $\pi/2$ \\ 
        2~-~2\_1    & $\theta_{2\_1}$ & 0       &$a_{2\_1}$ & 0 \\ 
        2\_1~-~2\_2 & $\theta_{2\_2}$ & 0       &$a_{2\_2}$ & 0 \\ 
        2\_2~-~3    & $\theta_{3}$    & 0       & 0         & $-\pi/2$ \\ 
        3~-~4       & 0               & $d_4$   & 0         & $-\pi/2$ \\ 
        4~-~5       & 0               & $d_5$   & 0         & $\pi/2$  \\ 
        5~-~6       & 0               & 0       & $a_{6}$        & $\pi/2$ \\  
        \bottomrule
    \end{tabular}
    \label{tab:DH}
    \vspace{-0.5cm}
\end{table}

\label{sec:problem}


\subsection{The Unified Teleoperation Control for Humanoid Robot}

 The upper limb of humanoid robots typically relies on existing teleoperation systems that are based on end-effector positioning. This approach limits the ability to control full-arm movements or requires the design of corresponding master control devices tailored to the specific robot, lacking general applicability. Inspired by our previous work, we have developed a relatively unified control method while also considering the issue of control precision. Essentially, this involves aligning the postures or positions of key joints, specifically the shoulder posture (humeral posture), elbow joint angle, wrist posture, and tracking the 6 DoF of the hand fingers. Therefore, the teleoperation control for humanoid robots can be divided into Shoulder Pose, Elbow Joint, Wrist Pose, and Hand Fingers Joint tracking controller. The core idea of the controllers is velocity (twist) tracking control with position (pose) closed-loop decreasing the motion tracking error, through the impedance control, where the active force components $\boldsymbol{F}_{ft}$ are introduced.

Due to the unique characteristics of the exoskeleton shoulder, we take shoulder posture tracking control as an example. As shown in \figref{fig:exo_control}, the four motors SC1-SH2 constitute the shoulder posture movement. Using the kinematic forward solution from the previous section, we can derive the three Euler angles for the humerus's directional movement speed~$\hat{\dot{\boldsymbol{q}}}^{m}$~and the quaternion pose~${\boldsymbol{q}}^{m}$.

 The shoulder state of the controlled humanoid robot can be represented by~${\boldsymbol{q}}^{s}$~and~$\hat{\dot{\boldsymbol{q}}}^{m}$.
The difference between them forms the target quantity of the impedance controller.
\begin{equation}
\begin{array}{l}
{\boldsymbol{q}}^{t} = {{\boldsymbol{q}}^{s}}^{*}{{\boldsymbol{q}}^{m}}, ~~\dot{\boldsymbol{q}}^{t} = \hat{\dot{\boldsymbol{q}}}^{m}-\hat{\dot{\boldsymbol{q}}}^{s}.\\
\end{array}
\label{equ:qt}
\end{equation}
where~${{\boldsymbol{q}}^{s}}^{*}$  represents the conjugate of the quaternion. Therefore, the control command variables of the shoulder motors of the humanoid robot can be obtained,
\begin{equation}
{\boldsymbol{\tau}}_{cmd}^{s} = \boldsymbol{k_p}f_{P}^{-1}({\boldsymbol{q}}^{t})+\boldsymbol{k_d}\boldsymbol{J}^{-1}\dot{\boldsymbol{q}}^{t} + \boldsymbol{\lambda}\boldsymbol{J}^\top\boldsymbol{F}_{ft},\\
\label{equ:tau_cmd}
\end{equation}
where, $\boldsymbol{k_p}$ and $\boldsymbol{k_d}$ denote stiffness and damping coefficients.  $f_{P}^{-1}$ represents the inverse kinematics function of the shoulder position layer. $\boldsymbol{J}$ is the Jacobian matrix of the shoulder movement of the humanoid robot. The item $\boldsymbol{\lambda}\boldsymbol{J}^\top\boldsymbol{F}_{ft}$ directly adds the influence of the interactive force $\boldsymbol{F}_{ft}$ between the user and the Exoskeleton to improve tracking sensitivity.
If that is for the tracking control of the elbow joint or each finger of the hand, \eqref{equ:qt} and \eqref{equ:tau_cmd} can be different,
\begin{equation}
\begin{array}{l}
{{q}}^{t}_{i} = {q}^{m}_{i}-{q}^{s}_{i}, ~~\dot{q}^{t} = \hat{\dot{q}}^{m}_{i}-\hat{\dot{q}}^{s}_{i}.\\
{\tau}_{cmd-i}^{s} = k_p{q}_{t}+k_d\dot{q}^{t}_{i} + \lambda\tau_{ft}.
\end{array}
\label{equ:qt2}
\end{equation}
where~${q}^{m}_{i}$,~${q}^{s}_{i}$~and~$\hat{\dot{q}}^{m}_{i}$, $\hat{\dot{q}}^{s}_{i}$ denote the position and velocity of each joint.
In addition to the control of the operated robot, the control of the master exoskeleton is also extremely important. Along with the dynamic compensation, we have added the active force control based on the full-arm coordination mechanism~(FCM) developed in our previous work~\cite{cheng2025flexibleexoskeletoncontrolbased},
\begin{equation}
\boldsymbol{{\tau}}_{cmd}^{m}=\boldsymbol{{\tau}}_{com}^{m}+\boldsymbol{{\tau}}_{h}^{m},
\end{equation}
where the six-dimensional force sensor data at the junction of the upper arm and forearm operates on the coordinated movement of the upper and lower arms, enhancing the movement flexibility and general applicability of the wearing of the exoskeleton.

Additionally, the control system software comprises four modular components: exoskeleton node, humanoid robot node, teleoperation controller node, and low-level driven program of the humanoid robot. By changing the low-level driven node of the humanoid robot, it can control another humanoid robot; by concurrently opening multiple identical control nodes, it can control several robots simultaneously.

\begin{figure}[t]
  \centering
   \includegraphics[width=1\linewidth]{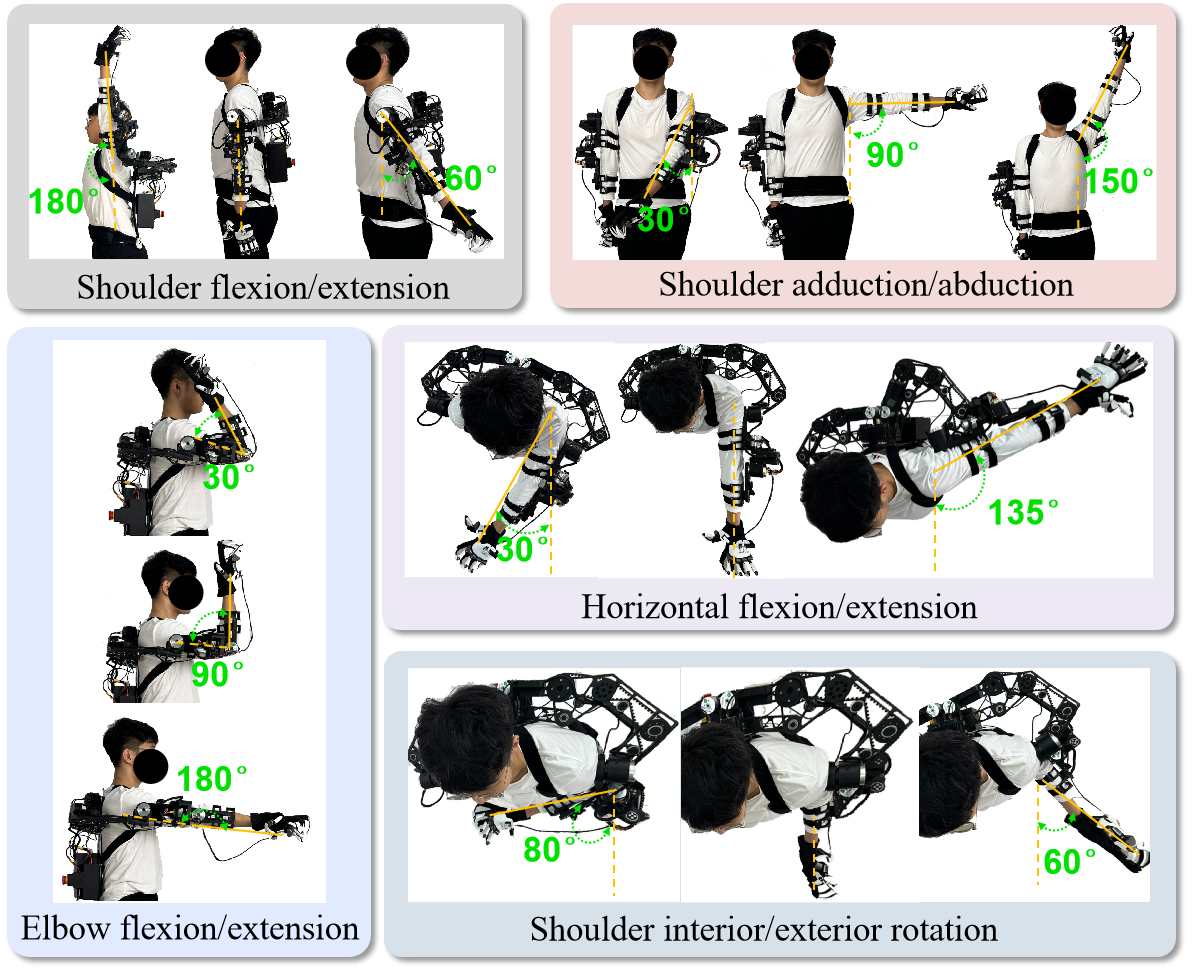}
    \caption{ROM of shoulder and elbow with the NuExo wearing.}
    \label{fig:exo_ROM}
    \vspace{-0.5cm}
\end{figure}

\section{EVALUATION AND EXPERIMENTS}
\label{sec:exp}
We conducted a series of experiments to evaluate NuExo’s performance across three critical dimensions: motion range coverage, pose collection accuracy, and real-world teleoperation effectiveness. The first experiment tested the maximum joint angles of NuExo, confirming its ability to replicate the full scope of human arm movements. In the second experiment, a direct comparison with inertial motion capture~(IMC) systems highlighted NuExo’s superior accuracy in tracking shoulder joint kinematics during dynamic motions. Finally, practical teleoperation trials involved multiple untrained operators controlling different humanoid robots through diverse manipulation tasks, demonstrating the system’s robust adaptability across platforms and scenarios. These results collectively validate NuExo’s ability to deliver precise, calibration-free control while maintaining intuitive operation—key advantages for real-world deployment.

\begin{figure}[tbp]
    \centering
    \begin{minipage}{1\linewidth}
    \includegraphics[width=1.00\linewidth]{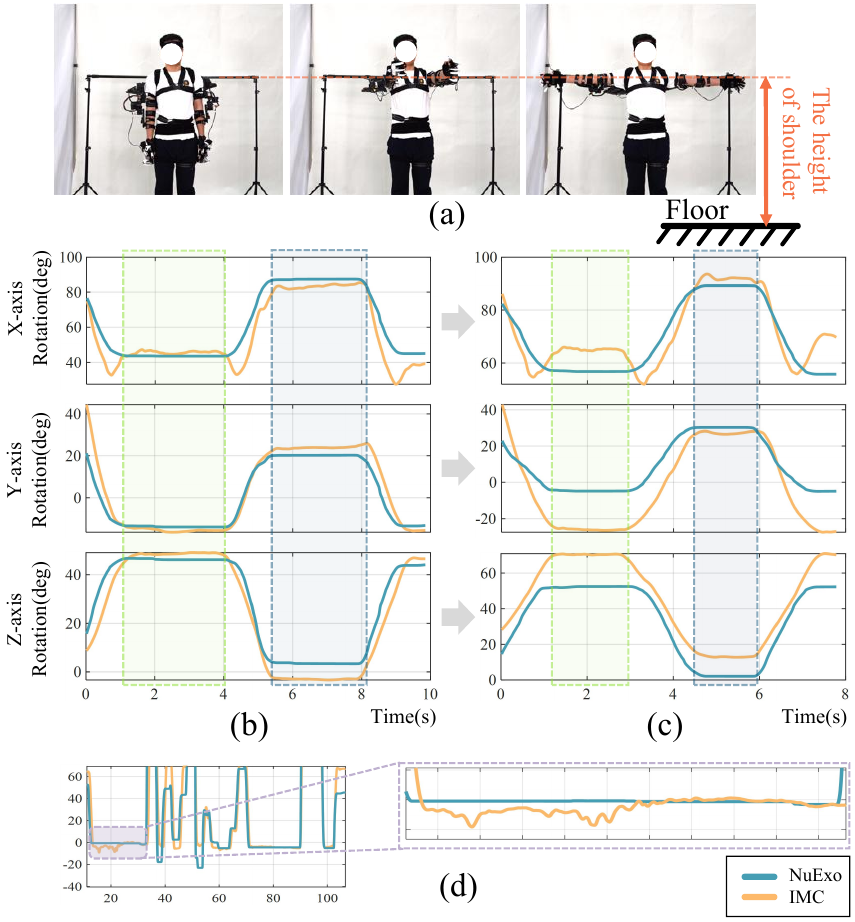}
    \vspace{-0.6cm}
    \caption{(a) Standard movements. (b) Comparison of shoulder joint posture data between two systems before perturbation. (c) Comparison after perturbation. (d) Comparison of data noise.}
    \vspace{0.2cm}
    \label{fig:mocap}
    \end{minipage}
    \begin{minipage}{1\linewidth}
    \centering
    \captionof{table}{Error Comparion~~{(Unit: rad)}}
    \footnotesize
    \setlength{\tabcolsep}{3pt}
    \begin{tabular}{cccccccc}
        \toprule
        &~& \multirow{3}*{\begin{tabular}[c]{@{}c@{}}Max.\\ deviation\end{tabular}} & \multirow{3}*{\begin{tabular}[c]{@{}c@{}}Avg.\\ deviation\end{tabular}} & \multicolumn{2}{c}{Static Max.} & \multicolumn{2}{c}{Static Avg.} \\ 
        &~& ~ &~ & \multicolumn{2}{c}{deviation} & \multicolumn{2}{c}{deviation} \\
        \cline{5-6} \cline{7-8}
        \vspace{-0.2cm}
        &~& ~ & ~ & ~ & ~  & ~  & ~  \\
        &~& ~ & ~ & EXO & IMC & EXO & IMC \\

        \midrule
        \multirow{3}{*}{Start} & x & 0.34 & 0.09 & \textbf{-0.05} & -0.14 & \textbf{-0.05} & -0.12 \\ 
        & y & 0.41 & 0.04 & \textbf{0.04} & 0.02 & \textbf{0.03} & 0.01 \\ 
        & z & 0.24 & 0.06 & \textbf{0.07} & -0.06 & \textbf{0.06} & -0.05 \\ 
        \multirow{3}*{\begin{tabular}[c]{@{}c@{}}After\\ perturbation\end{tabular}} & x & 0.26 & 0.20 & \textbf{0.13} & 0.23 & \textbf{0.07} & 0.17 \\ 
        & y & 0.39 & 0.21 & \textbf{-0.08} & -0.41 & \textbf{-0.07} & -0.37 \\ 
        & z & 0.33 & 0.22 & \textbf{0.04} & 0.31 & \textbf{0.03} & 0.26 \\
        \bottomrule
    \end{tabular}
    \label{tab:mocap data}
    \vspace{-0.5cm}    
    \end{minipage}
    \vspace{-0cm}
\end{figure}

\begin{figure*}[t]
\vspace{-0.2cm}
  \centering
   \includegraphics[width=1\linewidth]{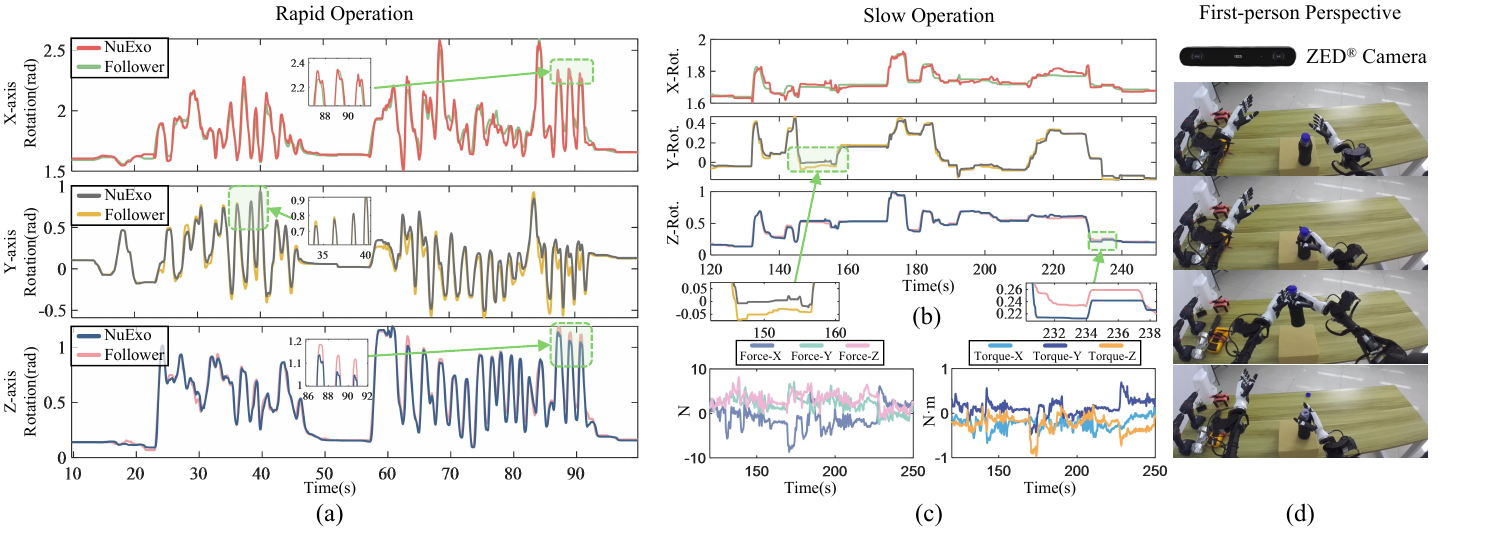}
    \captionof{figure}{(a) High-speed movement comparison: Real-time alignment between NuExo and the follower robot's left shoulder motions, confirming stable tracking during rapid operations. (b) Low-speed optimization: A filter pauses output during minor movements, effectively reducing hand tremors during delicate tasks. (c) Force data collection: Built-in sensors measure left-arm operational forces between the user and exoskeleton. (d) Immersive interface: The robot's first-person view streams to AR goggles, enabling intuitive real-time control.}
    \label{fig:exp3}
    \vspace{-0.5cm}
\end{figure*}

\subsection{Exoskeleton ROM Assessment}
To enable the NuExo system to replicate the native kinematics of the human GH joint, a new exoskeleton mechanism was developed in Section 4, and we conduct the ROM measurement experiment shown in \figref{fig:exo_ROM}. 
The shoulder can achieve 180° of flexion and 60° of extension, 30° of adduction and 150° of abduction. Users can easily raise their arms above their heads. Horizontal flexion and extension can reach 30° and 135° respectively. In the attached video, we demonstrated that it can still reach the above-mentioned ranges when performing circular motions.
The results demonstrate that the exoskeleton preserves substantial upper limb ROM when worn, including all of the natural arm movement patterns. This kinematic performance surpasses previous exoskeleton systems \cite{kim2017ijrr} and \cite{zimmermann2023tro}, confirming the efficacy of our structural design,  particularly the anthropomorphic shoulder mechanism emulating GH joint articulation.

\subsection{Comparison of Data Collection Accuracy with IMC}

IMC is one of the few devices that can provide untethered operation, extended mobility, portability, and multi-joint pose tracking, similar to our exoskeleton. Therefore, we conduct a comparison experiment on the data collection accuracy between NuExo and a commercial IMC system (Noitom® PN studio), as shown in \figref{fig:mocap} (a).
To ensure motion repeatability and set a ground truth, a steel frame aligned with the operator’s acromion height was positioned posteriorly, constraining arm elevation to a standardized plane. To ensure the consistency of movements, users wear both the IMC and the exoskeleton simultaneously.
The experimental workflow comprised four phases: \textbf{Calibration:} Both IMC and exoskeleton systems underwent calibration.
\textbf{Baseline acquisition:} The operator performed pre-set shoulder movements (\figref{fig:mocap} (b)).
\textbf{Dynamic perturbation:} The operator engaged in 10 minutes of unstructured high-intensity movements (e.g., running, jumping).
\textbf{Post-perturbation measurement:} Identical shoulder abduction tasks were repeated (\figref{fig:mocap} (c)).

Experimental data in Tab.~\ref{tab:mocap data} reveal that both IMC and NuExo achieve high initial accuracy post-calibration, with static mean errors approximating 0.1 rad. After high-intensity dynamic motions, IMC exhibited significant error amplification, reaching peak static errors of 0.41 rad, whereas NuExo maintained stable performance with static errors consistently below 0.14 rad. This discrepancy is attributed to the slippage of IMC’s strap-mounted IMUs during high-intensity movements, which disrupts sensor-body alignment and propagates errors through its whole-body kinematics model. In contrast, NuExo’s rigid linkage architecture directly measures relative joint angles via mechanical encoding, inherently isolating measurement integrity from motion-induced perturbations.

Further analysis of static noise characteristics (\figref{fig:mocap} (d)) demonstrates IMC’s elevated angular jitter compared to NuExo’s sub-noise performance. This phenomenon stems from error accumulation in IMC’s distributed IMU network, where individual sensor noise aggregates through forward kinematics computations. These results validate NuExo’s superior metrological robustness for wild deployment scenarios requiring sustained operational reliability.

\subsection{NuExo's Performance in Teleoperation Scenarios}

We implemented NuExo in practical teleoperation scenarios, with experimental results shown in \figref{fig:exp3}.
In high-dynamic motion tracking experiments, our controller demonstrated stable performance with an average angular error of 0.015 rad, while peak errors (0.05 to 0.08 rad, see zoomed view in \figref{fig:exp3} (a)) primarily arose from the follower robot's mechanical inertia during abrupt motion reversals. For slow-speed operations (\figref{fig:exp3} (b)), the average error further reduced to 0.01 rad. We designed a customized filter that suppresses human tremor below 0.015 rad, enhancing precision in delicate tasks.

The integrated force sensor on the left upper arm (\figref{fig:exp3} (c)) captured novel biomechanical interaction data unique to NuExo, offering potential advancements in imitation learning. Simultaneously, the ZED$^\circledR$ camera system (\figref{fig:exp3} (d)) provided dual functionality: delivering real-time stereoscopic feedback to the operator’s AR glasses for immersive control, and recording first-person visual datasets. Different from UMI~\cite{chi2024rss_UMI}, these synchronized datasets eliminate the need for joint coordinate calculations, enabling direct training of robotic systems using raw joint pose records.

\begin{figure}[t]
  \centering
   \includegraphics[width=0.95\linewidth]{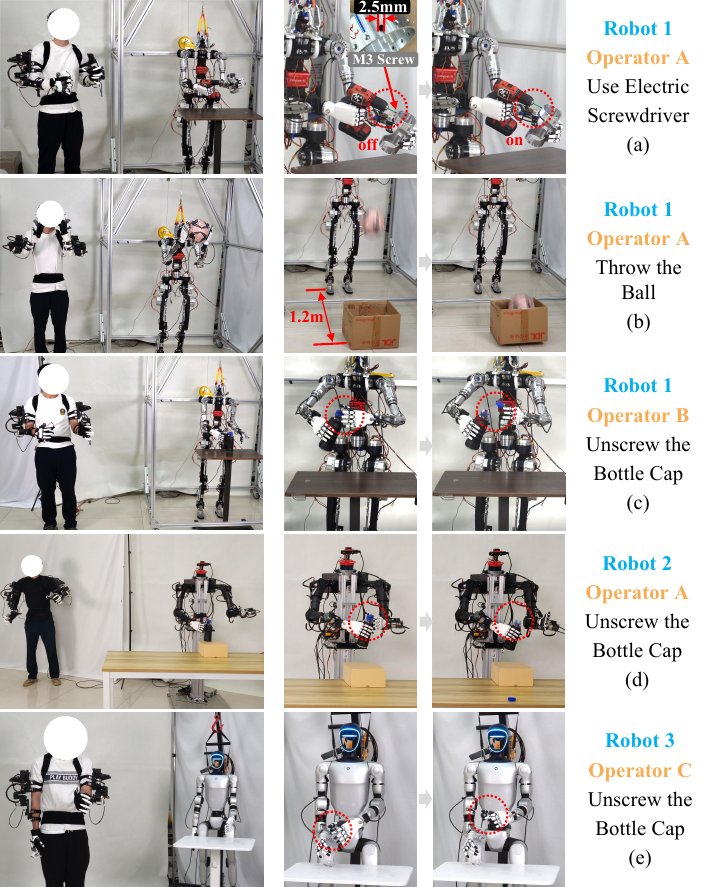}
    \caption{(a) and (b): The same operator using NuExo to control same robot to complete different tasks. (c), (d) and (e): Different operators controlling different robots to complete the same task.}
    \label{fig:/teleoperation_test}
    \vspace{-0.5cm}
\end{figure}

\figref{fig:/teleoperation_test} (a) illustrates the capability of NuExo system in fine manipulation tasks. With the NuExo, the operator can control the robot to use an electric screwdriver to tighten a screw (the diameter of the screw head is merely 2.5 mm). The operator is even able to control the switch of the electric screwdriver \textbf{with fingers}. In contrast, \figref{fig:/teleoperation_test} (b) demonstrates the ability of NuExo in high-dynamic tasks. The operator can maneuver the robot to accurately throw a ball into a box located 1.2 meters away.
We let different operators utilize the NuExo to control various robots for the completion of the same task. The operators could effortlessly finish the task without undergoing prior training (\figref{fig:/teleoperation_test} (c) to (e)).
NuExo enables direct utilization of the human arm, which is the most dexterous manipulator in nature, as a controller to guide robots through sophisticated operations.

\section{CONCLUSIONS}
\label{sec:conclusion}

In this paper, we present an innovative development: a lightweight, backpack-style exoskeleton robot, along with the construction of a generalized immersive humanoid robot teleoperation system and an upper limb multi-modal data collection system.
Our ingenious shoulder design has completely solved the problem of motor placement at the shoulder joints of active exoskeletons that consider shoulder anatomy. 
This enables the key concept (shoulder anatomical motion compensation) of the upper limb exoskeleton to be developed in a backpack-type design for the first time. According to the experiments, the ROM of the user's upper limbs and their daily movements remain unaffected when wearing NuExo. This verifies the comfort and the effectiveness of our exoskeleton design. Moreover, users can operate different humanoid robots to complete complex tasks without training, which validates the versatility and superior teleoperation ability of NuExo. Its data accuracy and long-term stability have been confirmed through operational data collection comparison, and our multi-modal sensing data (integrated force) is more comprehensive than that of the existing operation device shown in the related works.
In the future, we intend to explore the effectiveness of combining various types of data in the skill learning of humanoid robots.

\bibliographystyle{ieeetr}
\bibliography{nuexo}
\end{document}